\documentclass[11pt]{article}

\usepackage{graphicx}

\usepackage{deauthor,times}
\usepackage{xcolor}
\usepackage{url}
\usepackage{xspace}
\usepackage{wrapfig}
\usepackage{enumitem}
\setlist{nosep}
\usepackage{cite}
\usepackage{amsmath,amssymb,amsfonts}
\usepackage{algorithmic}
\usepackage{textcomp}
\usepackage{listings}
\usepackage{multicol}
\usepackage{hyperref}
\usepackage{cleveref}

\usepackage{textcomp}
\usepackage{tikz}
\usepackage{multirow, makecell, booktabs}
\usepackage{titlesec}
\usepackage{soul}

\definecolor{fgKeyword}{RGB}{0,92,184}   
\definecolor{fgRule}{RGB}{110,110,110}   
\definecolor{fgFrame}{RGB}{220,220,220}  

\setlist{leftmargin=*}

\newcommand{\code}[1]{{\text\small\texttt{#1}}}

\newcommand{\stitle}[1]{\smallskip\noindent\textbf{#1}}

\definecolor{blue}{HTML}{5383EC}
\newcommand{\red}[1]{\textcolor{red}{#1}\xspace}
\newcommand{\blue}[1]{\textcolor{blue}{#1}\xspace}

\definecolor{orange}{RGB}{230,126,34}   
\definecolor{teal}{RGB}{26,188,156}     
\newcommand{\orange}[1]{\textcolor{orange}{#1}\xspace}

\begin{document}

\title{Agentic Data Environments}
\author{
Elaine Ang,  Chenxi Huang,  Georgios Liargkovas,  Jerry Liu,  Jinhui Liu,  Nikos Pagonas,  \\
Charlie Summers,  Haonan Wang,  Jiakai Xu,  Tianle Zhou, 
Yusen Zhang,  \\
Zhou Yu,  Zhuo Zhang,  Tianyi Peng,  Kostis Kaffes,   Eugene Wu\\ 
  \href{https://dap.cs.columbia.edu}{\raisebox{-2pt}{\includegraphics[height=1em]{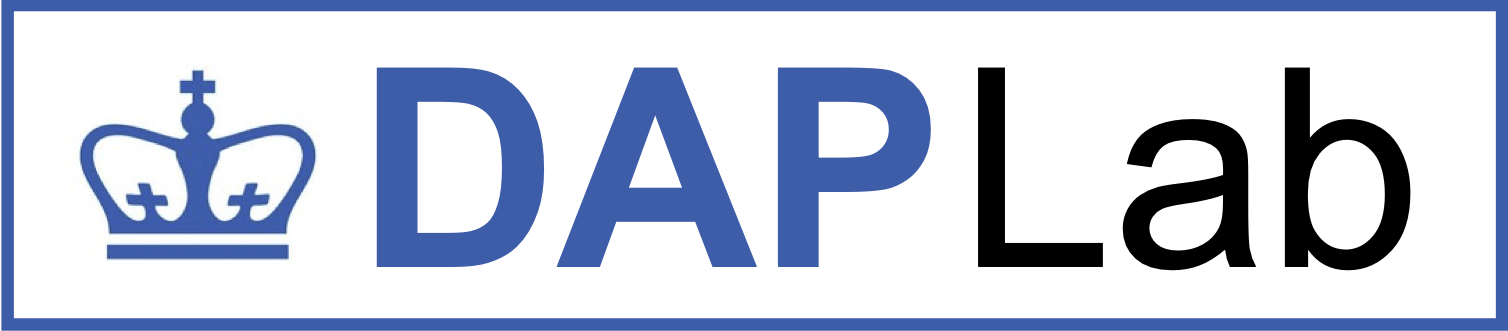}}}, Columbia University\\
  \small\{ra3448, ch4023, gl2902, jl6235, jl7309, np2948, cgs2161, hw2983, ax2155, mz2998, yz5296, \\
  \small zy2461, zz3474, tp2845\}@columbia.edu,  \{kkaffes,  ewu\}@cs.columbia.edu
}

\maketitle


\section{Introduction}

Automation has long been the promise of computing.  The introduction of modern large language models~\cite{Vaswani2017AttentionIA} (LLMs) has changed who (or what) performs this automation. LLMs, combined with vibe coding, agent frameworks, and rich API ecosystems, empowered non-programmers to deploy autonomous agents that operate terminals~\cite{Merrill2026TerminalBenchBA}, call APIs and tools~\cite{anthropic2024mcp,anthropic2025claudeskills,google2025a2a}, navigate GUIs~\cite{Xie2024OSWorldBM}, code~\cite{Suris2023ViperGPTVI}, and query databases~\cite{Li2023CanLA,Lei2024Spider2E}.  Rather than copilots that recommend actions for the user, agents autonomously observe data, plan and execute actions, and observe their {\it effects}.
{\it This shift from reading data to acting on it is the central challenge in future data management.}


Today's data agents are largely read-only. NL2SQL, retrieval-augmented question answering, and data analytics agents observe data, synthesize it, and return an answer. A tax reporting agent may retrieve financial statements and transaction records to estimate last quarter's revenue; its actions make no long-term side effects to the environment.
This design simplifies evaluation, improves failure tolerance, and limits potential harm.

In contrast, agentic automation mutates the environment with real consequences. The same tax scenario is fundamentally different when the agent also reconciles discrepancies across financial statements, applies tax logic, and files official returns. Each step is simultaneously a data write and a consequential action that e.g., modifies accounting records, overwrites prior filings, and submits legally binding documents. Because mutation and consequence are coupled, errors are not merely wrong answers: they can lead to regulatory penalties, lawsuits, or compliance violations. Agentic automation is ultimately a read-write problem: when agents can modify data, the value of automation shifts from what agents can accomplish to what happens when they fail.

\subsection{Automation's Value Proposition}

To make this trade-off precise, consider the core value proposition for agentic automation:
\begin{align}
\text{Value} = \text{Benefits} - \text{Costs}
\end{align}
Automation promises substantial benefits through speed, scale, and labor savings. However, the cost of failure differs in character and magnitude. Benefits accumulate gradually across many successes, but costs are abrupt, catastrophic, and difficult to reverse:  deleting a production database~\cite{pcmag_replit_vibe_coding_2025}, triggering a cloud outage~\cite{guardian_aws_ai_outage_2026}, and exfiltrating data~\cite{forbes_samsung_chatgpt_ban_2023,verge_chatgpt_gmail_shadow_leak,register_salesforce_prompt_injection_2025,codeintegrity_notion_ai_security}. Because agents operate over systems of record, failures can propagate before detection. In both perception and practice, the potential costs of agent automation therefore appear unbounded.

This asymmetry shapes adoption. Users do not calibrate trust based on overall performance, and a single salient failure can suppress adoption out of proportion to its likelihood~\cite{parasuraman1997humans}. This is corroborated by prospect theory, which finds that humans weigh losses much more heavily than the same gains~\cite{kahneman1979prospect}. As a result, those evaluating automation focus on worst-case outcomes rather than expected performance, and systems that are only safe in the common case remain unsuitable for important tasks.

The implication is that ``best-effort'' safety is not enough, as higher average reliability does not affect adoption if catastrophic outcomes remain plausible. We need to both increase the benefits of automation {\it and } bound the consequences of failure.  This is not solely an agent design problem, but a systems problem: the environment that the agent executes within and the guarantees the environment provides.  

\subsection{From Databases to Data Environments}

Databases remain a central component of modern computing. However, automation goes well beyond simply querying databases or perform analytics. Real-world tasks require interacting with the broader computing stack, including applications, APIs, files, configuration systems, command-line tools, and external services.

\begin{example}\it
An agent send an HTTP request to a web service that triggers server-side business logic, which  launches background jobs, calls external APIs, writes to the file system, and mutates the database.  Through this chain, the agent indirectly interacts with multiple subsystems and a considerable amount of evolving state: application variables in the server process, configuration files, job queues and logs produced by background tasks, files written to disk, responses from external services, and persistent records in the database.
The outcome and effects depend on database contents {\it as well as} the configuration and state of the surrounding environment.
\end{example}
This suggests that ``data management'' for agentic automation must extend beyond databases to the  \emph{data environment}: the collection of heterogeneous resources that the agent runs and interacts within---including data lakes, file systems, memory, APIs, derived artifacts, processes, and system metadata—along with the mechanisms that govern how this state is accessed, modified, and allowed to flow.  Unlike a classic Database Management System (DBMS), data environments encompass a broad range of data models and software components rather than a single data store. Unlike a dataspace~\cite{Halevy2006PrinciplesOD}, which focuses on integrating heterogeneous data sources, data environments are the \emph{stateful substrate} that the agent executes within.

\subsection{Towards Agentic Data Environments}

\begin{center}
{\it What does a data environment designed for agents rather than humans look like?}
\end{center}

\begin{wrapfigure}{r}{0.50\textwidth}
\vspace{-1.5em}
\centering
\includegraphics[width=\linewidth]{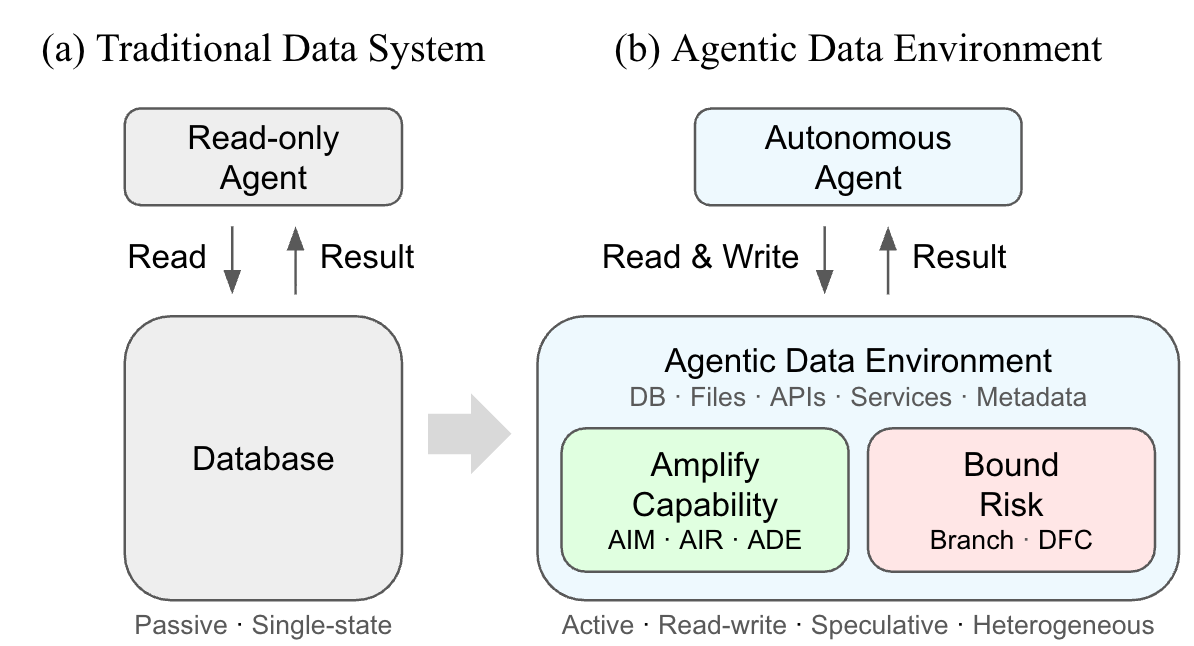}
\caption{This paper describes a shift from systems for today's analytic agents to Agentic Data Environments.}
\label{fig:shift}
\vspace{-.5em}
\end{wrapfigure}

Modern computing systems can already be viewed as a form of \emph{data environment}. Filesystems, databases, services, and APIs collectively form a shared state space within which programs operate. However, these environments are largely \emph{passive}: they serve data and compute requests, but do not actively organize information, guide decision making, or constrain data flows and data is used.

Agentic automation demands more. Agents continuously observe, act, and adapt to the data environment, and are capable of reasoning over large amounts of technical content far beyond any human. Yet, agents are ultimately consumers and producers of data: its decisions depend on the quality of the information it uses and its ability to safely explore its action space.
Thus, supporting agents requires the data environment itself to become more than a passive store of state.  These \textbf{\textit{Agentic Data Environments}} actively prepare information and govern agent interactions in order to support reliable and safe automation. Their role is to:
\begin{itemize}[itemsep=-.5em]
    \item 
\stitle{Amplify Capability.} the environment must actively discover, materialise, organize and expose information from heterogeneous sources---documents, databases, data lakes, APIs, and the system itself---in forms that empower the agent without requiring developers to manually engineer every data pipeline.
\item  \stitle{Bound Risk.} the environment must provide guarantees that today's computing stacks lack: isolation and sandboxing to limit failures, transactional semantics and versioning across components, branching and rollback for speculative exploration, and policy enforcement to control how agents use and transform information.
\end{itemize}
\smallskip
\noindent This paper outlines the foundations of \emph{Agentic Data Environments}. We describe three information management challenges based on how information becomes available: discovering agent-relevant data in massive data lakes, transforming data into agent-ready representations, and collecting/generating latent signals from the environment. We then discuss the infrastructure needed for safe agent exploration, including branching and data safety mechanisms.  These same environment capabilities are also critical for training better agents (\Cref{s:conclusions}).

\section{Agentic Data Environments To Improve Agent Capabilities}\label{s:benefits}


To increase the benefits of automation, agents must access and reason over the right information.
Agent failures are increasingly information rather than reasoning failures, and even the most powerful models cannot solve tasks when the relevant signals are missing, poorly structured, or undiscoverable.
%
\begin{center}
   {\it The responsibility of the data environment 
   is to manage, find, elicit, and deliver \\
   the right information, in the right representation, 
   at the right time, for the right task. }
\end{center}

This need is particularly acute for a new class of agent builders: domain experts who can vibe code agents without formal software development training. While they can identify relevant data sources or share domain knowledge, turning that information into agent-ready representations remains out of reach. Ultimately, the challenge is developer ergonomics: how to use domain expertise while hiding the complexity of data management?

The key observation is that the data consumer is no longer a human. Traditional data products aim to faithfully represent reality for analysts. Agents instead treat data as a means to an end: task success. Data environments must shift from representing reality toward preparing task-relevant signals for agents.

\begin{wrapfigure}{r}{0.5\columnwidth}
    \vspace{-.75em}%
    \includegraphics[width=.9\linewidth]{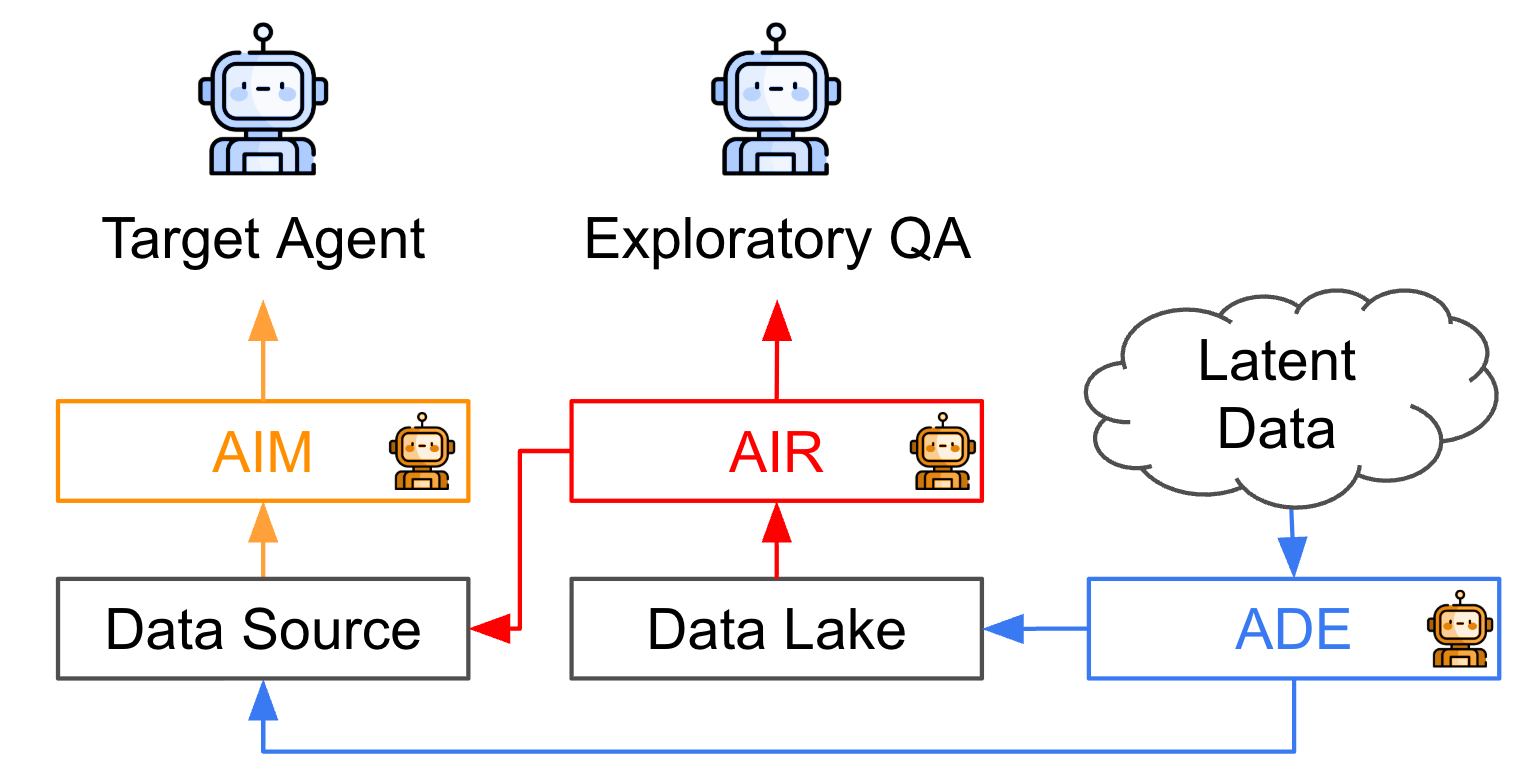}
    \vspace{-1em}
    \caption{\red{AIR} finds relevant data sources for EQA or target agents, \textcolor{orange}{AIM} turns sources into capabilities, and \blue{ADE} materializes missing signals into new data sources. \includegraphics[height=1em]{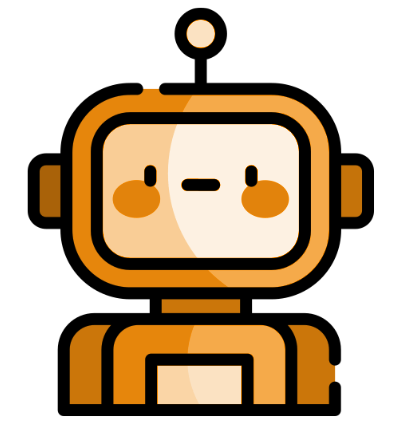} curates information to improve \includegraphics[height=1em]{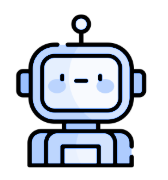}. }
    \label{fig:aim}
    \vspace{-.5em}
\end{wrapfigure}

The main variable across settings is how available the required information already is. 
In some cases the relevant data source is known but poorly structured for the task. 
In others the information exists but must be discovered within a large heterogeneous data lake. 
Finally, the needed signal may exist only implicitly in the environment and must be inferred or materialized.

These settings motivate three complementary directions. 
\orange{\emph{Agentic Information Management (AIM)}} transforms known sources into agent-ready capabilities. 
\red{\emph{Agentic Information Retrieval (AIR)}} finds sources with the needed evidence by a task. 
\blue{\emph{Agentic Data Elicitation (ADE)}} surfaces latent signals that are not yet materialized as artifacts.

\subsection{Agentic Information Management: Taking AIM at Data}
\label{sec:aim}

Today, the dominant approaches that make data accessible to agents will convert raw sources into vector embeddings for retrieval-augmented generation (RAG)~\cite{longmemeval}, or store them in a default log system~\cite{chhikara2025mem0}, document store~\cite{zhang2025agentic}, or text files~\cite{anthropic2025effective}. Many refer to such processes as agent context or agent memory curation. These generic representations impose a fixed schema, ignorant of the agent's task, and lose the structure that may be critical for downstream reasoning. For instance, using RAG for a conversation corpus loses temporal ordering, speaker identity, and cross-session relationships needed by a question-answering agent.

Tasks, data, and models evolve over time, and static representations quickly grow stale. Thus, we propose \textbf{\textit{Agentic Information Management (AIM)}} to automatically manage information representations to benefit the domain expert's target agent. Given a data source and high-level expert guidance, AIM proposes a data model, schema, and extraction pipeline that captures the structure it believes will best support the target agent's tasks. 

\begin{example}\it
LoCoMo~\cite{locomo} is a multi-session conversation dataset where two speakers (e.g., Caroline and Melanie) converse across 19~sessions over several months. The dialogues cover events, career plans, hobbies, family activities, and evolving personal interests. Rather than embed this corpus into a vector store, an AIM agent reads the dialogues and proposes a relational schema tailored for what it expects to need. For example, the agent may design tables for \texttt{Users}, \texttt{Sessions}, \texttt{Messages}, \texttt{Events}, \texttt{Interests\_Activities}, and \texttt{Relationships}. It then extracts and loads the session facts (e.g., {\it ``Caroline attended an LGBTQ support group on May~7, 2023''}) into the database, and exposes a SQL skill to this database.
At query time, the target agent simply writes SQL queries against this database. To answer {\it ``What hobby does Melanie use to relax?''}, it queries \texttt{Interests\_Activities} filtered by activity type rather than filtering hundreds of raw dialogues.  
\end{example}

\begin{wrapfigure}{r}{0.5\columnwidth}
    \vspace{-.75em}
    \centering
    \includegraphics[width=.95\linewidth]{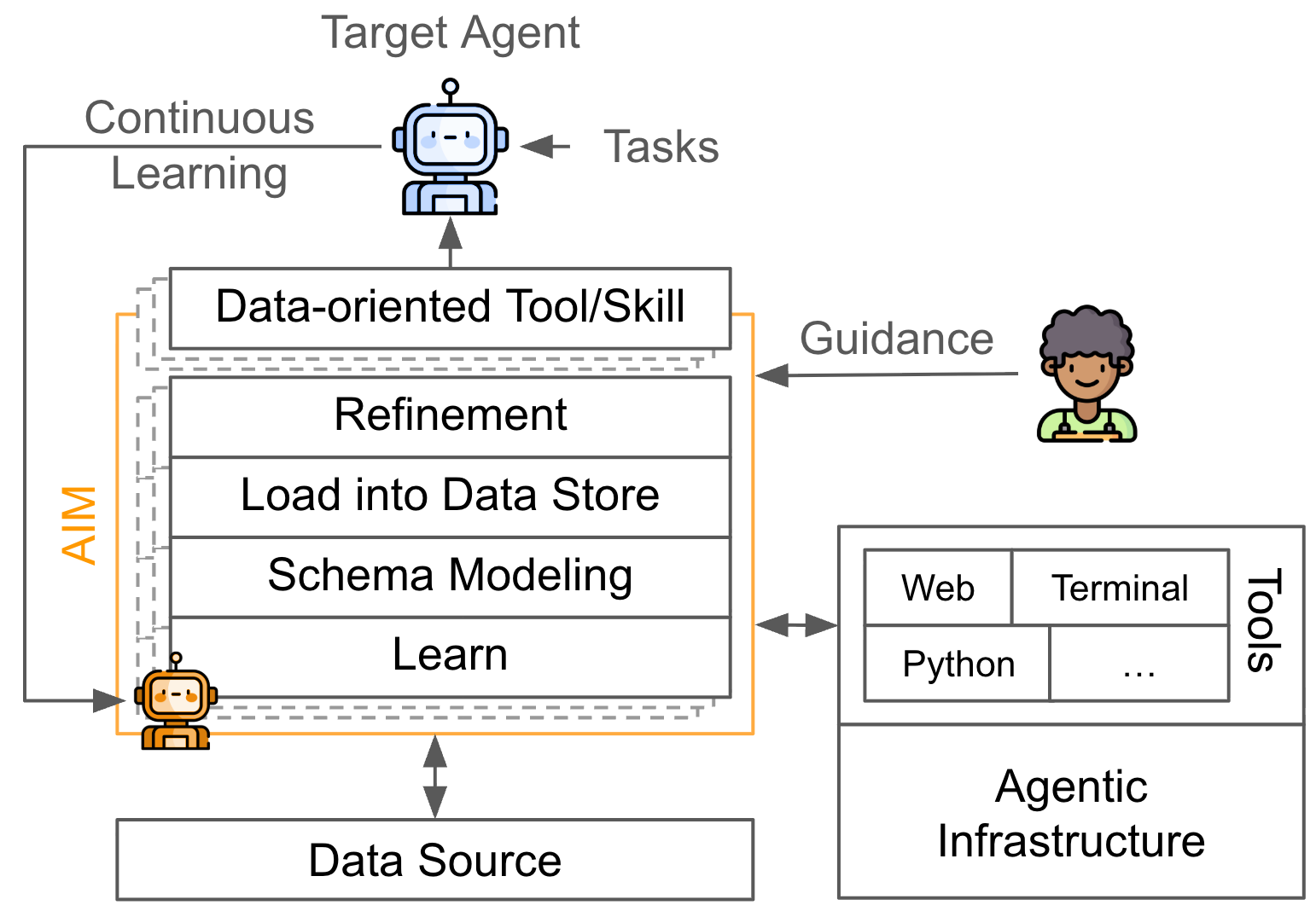}
    \caption{AIM takes a target agent, guidance, a set of tools, and a data source as input, and generates pipelines and artifacts to improve the target agent's task quality. }
    \label{fig:aim}
\end{wrapfigure}

\noindent
Concretely, AIM is a multi-agent system that progressively adds structure and task-specialization while preserving the ability to evolve (\Cref{fig:aim}).
The {\it Learning} stage analyzes the data source and hypothesizes data models for the target use cases.    {\it Schema Modeling} then instantiates schemas and constraints in a specific data system (e.g., RAG, RDBMS, etc). 
{\it Data Loading} generates an Extract-Transform-Load pipeline to load the data source into the data system, and {\it Refinement} performs physical design to generate views and indexes for fast access.
The resulting database is exposed as a \emph{Data-Oriented Tool} or \emph{Skill} that the target agent can call to retrieve task-relevant information.  
Each stage is informed by the domain expert's high level guidance (e.g., ``the last user should be useful'') and can use an extensible set of tools that run atop the Agentic Data Environment infrastructure (\Cref{s:costs}).
The pipeline is agentically generated, so stages are revised based on new guidance, task feedback, and source changes. 


\begin{example}\it
    On LoCoMo~\cite{locomo}, AIM's accuracy is comparable to adding the full dialogue to the context of a frontier model, but uses only ${\sim}10\%$ the context length. When compared with specialized agent memory systems like Mem0~\cite{chhikara2025mem0} and the SOTA RAG-based Octen~\cite{octen2025rteb}, AIM is $49.8\%$ and $15.82\%$ more accurate, respectively. 
    AIM is $4.18\times$ faster than the state of the art agentic memory system GAM~\cite{gam} and has $13.54\%$ higher relative accuracy on average.   In particular, AIM is $11.53\%$ more accurate for temporal and $24\%$ for open-domain reasoning questions because its structured representations are more effective than text file and RAG representations.
\end{example}

\subsubsection{Research Directions}
The similarity between AIM and an automated data integration pipeline is intentional. The key difference is the objective: rather than create a high-quality schema upfront, AIM rapidly bootstraps an \emph{agent-consumable capability} and evolves it over time.
However, using agents to manage schemas and representations introduces challenges. There is no ``correct'' schema: the same conversation be modeled as an event timeline for one task and a relationship  graph for another, and different extraction strategies can vary in performance. Further, the optimal representation is not static---models, prompts, and tools regularly change---so AIM must continuously ``sample'' for better pipelines (\textcolor{gray}{dashed gray boxes}) while accounting for migration costs.
%
Consequently, the data environment must be designed to rapidly generate, evaluate, and migrate pipelines; use task outcomes as feedback to improve or discard pipelines; and select the appropriate pipeline per request. 

\subsection{Agentic Information Retrieval: Breathing AIR into the Lake}
\label{sec:lake}
AIM assumes that the data has been found. In practice, useful information is distributed across heterogeneous collections of millions of documents and datasets---a \emph{data lake}---and the agent must first discover which sources contain the evidence needed to complete a task. We call this problem \textbf{\emph{Agentic Information Retrieval} (AIR)}.

\emph{Data discovery} systems seek to find datasets relevant to a query or example table. However, their evaluation measures---e.g., keyword matching, dataset similarity, joinability, or prediction accuracy---are only proxies for the downstream tasks users want. One common task is question answering (QA), where the goal is to retrieve evidence and synthesize an answer. QA spans reading comprehension~\cite{rajpurkar2018know, kwiatkowski2019natural, khashabi2018looking, dua2019drop}, open-domain question answering~\cite{joshi2017triviaqa, karpukhin2020dense}, and tabular question answering~\cite{pasupat2015compositional, nan2022fetaqa, herzig2021open}.

Recent LLM-based QA agents assume that the relevant evidence is in the context or easily found in a small curated corpus. In data lakes, however, the structure, semantics, and source relationships are largely unknown, and their scale is too large to fully analyze with LLMs. We call this setting \textbf{\emph{Exploratory Question Answering (EQA)}}: the agent must iteratively infer needed evidence and search the lake to find sources with the evidence.

\begin{wrapfigure}{r}{0.5\columnwidth}
    \vspace{-1em}
    \includegraphics[width=.925\linewidth]{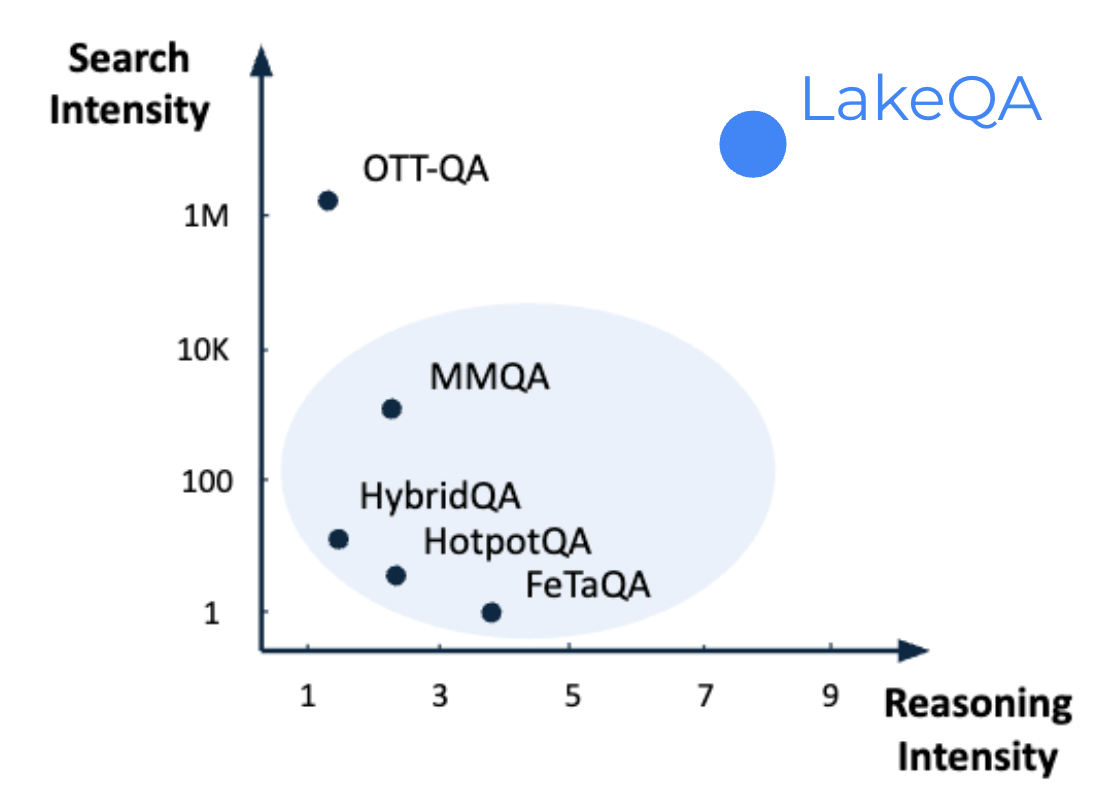}
    \caption{LakeQA is the first exploratory QA benchmark that pushes search intensity and reasoning intensity.  Its $1000+$ tasks are created and validated by 4 independent annotators including one database PhD.  }
    \label{fig:lakeqa}
    \vspace{-.5em}
\end{wrapfigure}

To study this, we constructed \textsc{LakeQA}, an EQA benchmark over a 9.5~TB data lake containing ${\sim}40M$ documents from Wikipedia and Data.gov. Tasks must find and reason across multiple heterogeneous sources (7.67 documents on average) drawn from millions of candidates.
Of particular importance is that tasks must {\it require search over the data lake}.   

In many tasks, LLMs can easily hallucinate correct answers to individual steps or the entire task.   To enforce this property, each step 1) must require accurate answers from previous steps to formulate its evidential need, and 2) the answer must be data dependent on a derived statistic or inferred fact from content in the lake.   
For example, identifying neighborhood schools that satisfy a class-size constraint requires locating datasets describing school statistics and neighborhood boundaries, and reasoning across them sequentially.

To ensure that the $1000+$ tasks are of high quality and do not contain annotation errors~\cite{Jin2026PervasiveAE}, the tasks are created by a team of 5 database Ph.D. students and 4 senior computer science undergraduates that passed our data science proficiency exam.  Each task is validated by 4 annotators including at least one Ph.D.

LakeQA uniquely stresses two dimensions.  \emph{Search intensity} measures the difficulty to find relevant sources in the data lake, and \emph{Reasoning intensity} measures the amount of multi-document inference required given those sources.  
As shown in \Cref{fig:lakeqa}, existing benchmarks emphasize only one axis: multi-hop QA increases reasoning depth but provides the relevant documents, while dataset search emphasizes discovery but does not require downstream reasoning. 
In contrast, LakeQA requires both.

Across seven frontier models, end-to-end accuracy is ${\leq}23\%$. The dominant failure mode is failure to find the required datasets, not reasoning.   This suggests that EQA agents must explore the data lake as part of the reasoning process; at this scale, the search system's design is as critical as the model's capability.

\subsubsection{Research Directions}
\textsc{LakeQA} shows that AIR must address two related challenges: represent the contents of a massive data lake so relevant sources can be discovered efficiently, and reason over how those
sources can be combined to produce an answer.
Discovery motivates a \emph{semantic layer} between the agent and the data lake that summarizes what data exists, what it represents, and how sources may relate. Such summaries necessarily compress the lake. If compressed too aggressively, relevant sources become undiscoverable; if too coarse, the agent must evaluate large numbers of false positives that quickly exhaust latency, token, and cost budgets.
Reasoning challenges arise when candidate sources are found. Answering a question must compose evidence across datasets by applying e.g., joins, entity resolution, temporal alignment, or reconciling conflicting definitions. The space of possible compositions is combinatorial, so AIR must jointly reason about \emph{what sources to retrieve} and \emph{how to compose them} to allocated limited compute and context to the most promising reasoning paths.

\subsection{Agentic Data Elicitation: Distilling ADE from the Ether}
\label{sec:latent}
The previous subsections assume that the relevant information already exists as data and must only be found or organized. 
In practice, however, many tasks depend on information that is only implicitly present in the environment.
This \emph{latent data} refers to signals that can be surfaced from the environment through observation, synthesis, or controlled experimentation. 
\emph{Agentic Data Elicitation (ADE)} turns these implicit signals into explicit artifacts that future agents can reuse.

Latent data may capture semantic structure---such as implicit table roles, undocumented relationships between attributes, or derived business logic---or performance structure---such as workload regimes, bottleneck signatures, and relationships between control knobs and system metrics.
%
%
The agent decides what to inspect, hypotheses to test, experiments to run, and candidate artifacts to retain.
The environment exposes the observation and control surfaces that make this possible---e.g., example, metadata, row samples, safe program execution, workload traces, metric probes, and actuation interfaces---and provides mechanisms to validate, store, and serve the resulting artifacts. 
Elicitation may be \emph{passive}, when the agent collects existing state or history, or \emph{active}, when it performs controlled interventions to reveal hidden response structure.

\begin{example}\it
Consider a simple example from NL2SQL systems. A user asks: ``How many sales activities does each account have?'' The database contains tables \texttt{ActivityHistory}, \texttt{Task}, and \texttt{Event}. An agent that relies only on schema names may incorrectly treat \texttt{ActivityHistory} as the fact table. In practice, business activities are stored in \texttt{Task} and \texttt{Event}, while \texttt{ActivityHistory} records change logs. 

Through ADE—inspecting schema structure, sampling rows, or observing prior query traces—the agent can elicit the latent artifact ``\texttt{ActivityHistory} is a change-log table.'' Once materialized, this artifact becomes another AIM object that improves downstream reasoning such as table selection and query synthesis.
\end{example}
Various agents interacting with systems already follow this pattern.
The Tk-Boost~\cite{tkboost} NL2SQL system materializes ``tribal knowledge'' by generating corrections from past interactions.
REDSQL~\cite{redsql} materializes latent constraints (e.g., join consistency, valid aggregations) not in the schema; 
AgentSM~\cite{agentsm} materializes example trajectories by synthesizing question sets and multi-step reasoning traces. 
This perspective extends from semantics to performance, where the objective is optimization rather than correctness.
For example, to tune an OS scheduler~\cite{liargkovas2025expertinresidence}, the agent must elicit how scheduler knobs affect performance by collecting runtime measurements, system state, and domain hints.
Although their mechanisms differ, these approaches distill implicit structure from experience, constraints, or reasoning traces into explicit artifacts that persist beyond a single task.


Therefore, ADE-capable environments (e.g., OS, DBMS, system components) should expose APIs for agents to probe and learn from implicit signals. Beyond passive observation, environments should also provide safe mechanisms for agents to run controlled experiments (e.g., testing alternative indexes, cache policies, or operator strategies) while ensuring that changes remain sandboxed, auditable, and reversible. 
In a sense, ADE capabilities resemble eBPF-style extensibility~\cite{sodhi2025ebpfml,zussman2025cacheext}, where agents attach lightweight logic to system hooks to measure behavior, test hypotheses, and materialize reusable artifacts about system dynamics.

Many useful artifacts follow this pattern.
In data-centric settings, agents may elicit statistical summaries, common join paths, derived metrics embedded in business logic, or schema groupings learned from past workloads.
In systems settings, agents may elicit bottleneck signatures, knob-sensitivity maps, workload phases, or regime shifts.
These artifacts improve performance not by changing the underlying model, but by making more of the environment's structure explicit and available to future agents.

\subsubsection{Research Directions}
This opportunity is to reason about latent structure and materialize it. 
Once elicited and validated, many artifacts---such as performance counters, inferred schema relationships, reusable query hints, or tuning knowledge---can be managed as standard AIM data sources and reused across tasks.
To support this ADE-capability, environments must make latent structure easy to elicit, validate, persist, and refresh. 
This demands low-latency access to instrumentation and sampling interfaces,
methods to validate potentially spurious or overfit candidate artifacts before they are reused,
efficient ways to maintain these artifacts an environments (e.g., schemas, business logic, tasks) evolve,
and lifecycle management of these artifacts.

%
\section{Agentic Data Environments for Safe Agent Automation}\label{s:costs}

Autonomous agents place new demands on data environments. While \Cref{s:benefits} focused on preparing information to improve agent capabilities, this section addresses the complementary challenge: enabling agents to explore real systems while bounding the risks of their actions. 

\begin{center}
{\it The responsibility of the data environment is to let agents explore aggressively \\
while ensuring that environment state and data remain protected.}
\end{center}

Agent exploration is the ability to trial and error, call tools, mutate data, and revise plans.  These trials may modify databases, system state and configurations, or external services.  Such exploration must not corrupt shared state or trigger irreversible side effects.
At the same time, freedom to explore is not sufficient. An agent that can freely read sensitive data, combine signals in ways that violate policies, or exfiltrate results to external services is unsafe regardless of how cleanly its exploratory state is managed.  

To address these risks, data environments must enforce two complementary properties. 
\textbf{Branching} protects \textbf{\emph{state safety}} by allowing agents to interact with live environments in isolated speculative copies. \textbf{Data Flow Control} (DFC) protects \textbf{\emph{data safety}} by constraining how information may propagate through the system: from sources such as databases, files, and retrieval stores to sinks such as tables, prompts, tools, or external APIs.

Together, these mechanisms let agents explore while maintaining desired safety guarantees.   Branching contains the agent’s actions within isolated speculative environments, while DFC governs which data may enter or leave those environments. By decoupling agent
capabilities from safety guarantees, agents can automate within the data environment without requiring developers to reason about every possible agent trajectory.

\subsection{Branching for Agent Exploration}

Autonomous agents rarely solve complex tasks through a single linear execution. Instead they must explore alternative trajectories and compare intermediate outcomes. Recent agent frameworks therefore incorporate search mechanisms such as reflection, hierarchical planning, or tree search, and empirical studies show that enabling exploration over intermediate states substantially improves success rates on long-horizon tasks \cite{xu2025systemsfoundationsagenticexploration}.

Exploration requires systems to branch and restore from arbitrary intermediate states. Unlike traditional workloads, an agent’s state includes intermediate outputs as well as the state throughout the data environment.

\subsubsection{Branchable DBMSes}
Databases already maintain multiple logical versions of data: MVCC snapshots and savepoints let transactions observe consistent states and perform limited rollback.
More recently, branchable DBMSes advertise primitives to {\it branch} database state. Examples include WAL-based approaches (e.g., Neon) and content-addressed storage engines (e.g., Dolt), where branches are lightweight pointers into shared storage structures. These metadata-level mechanisms enable isolated experimentation without expensive data copying.

However, these architectures assume a few long-lived branches created by humans.
Agent exploration instead generates hundreds or thousands of short-lived states.
Further, mutations in a branch may be logical (e.g., updating data or modifying schemas) or physical (e.g., creating indexes, materialized structures, or changing system configurations), and evaluation may read data within a branch or across many branches.  

To better understand the above, we built \emph{BranchBench}, a benchmark that models the core exploration pattern:  a repeated \emph{branch–mutate–evaluate} loop. For example, an agent optimizing database performance may repeatedly branch the current state, apply a candidate modification (e.g., index creation or layout change), run evaluation queries, and either discard or extend the branch.   

The benchmark simulates five agentic applications that span exploration depth and fanout, mutation intensity (logical and physical), and branch life-cycle management.  
{\it Software engineering} and {\it Failure reproduction} emphasize rapid branch creation and high-throughput execution. {\it Data curation} stresses cross-branch analytics and comparison, while  {\it MCTS} stresses dynamically shaped exploration trees and many active branches.   {\it Simulation for planning} generates wide bursts of short-lived branches. 
These workloads demand that branches must be created quickly, mutations must be isolated, queries must execute efficiently within each branch, cross-branch queries must be efficient, and branching must be storage- and resource-efficient. 

On Neon, Xata, Tiger, Dolt, and PostgreSQL, we find that today's \textbf{branchable databases do not support agentic workloads.}
No system successfully completed the benchmark, even at a modest scale factor. 
Branch-optimized systems saw read query latencies degrade by $5{–}4000\times$ depending on branch count and concurrency. Conversely, query-optimized systems incurred $25{–}1500\times$ higher branch creation latency. None efficiently query across branches, and many took seconds or minutes to allocate branches.

The key reason is that \textbf{agentic exploration is more aggressive than classic branching.} While developers only create a few long-lived branches, techniques such as Monte Carlo Tree Search require hundreds or thousands of speculative states. In a $1000$-step MCTS experiment, Neon completed $3\%$ of steps due to limits on concurrent branches, while DoltgreSQL only completed $17\%$ by the 2 hour timeout because reads degrade with the number of branches.    This performance  trades off with resource overheads. For instance, Neon allocates dedicated compute instances for each branch and required $43\times$ more storage usage under schema-heavy workloads.  
It is clear that a \emph{branch-native DBMS} designed for high-frequency speculation is needed.

\subsubsection{Branching Beyond the DBMS}
Database branching is necessary but not sufficient---real agents are not confined to the DBMS.
They use the database, filesystem, process memory, terminal context, caches, application runtimes, and sometimes external services.
The data environment must natively support branching.

\begin{wrapfigure}{r}{0.45\columnwidth}
    \includegraphics[width=\linewidth]{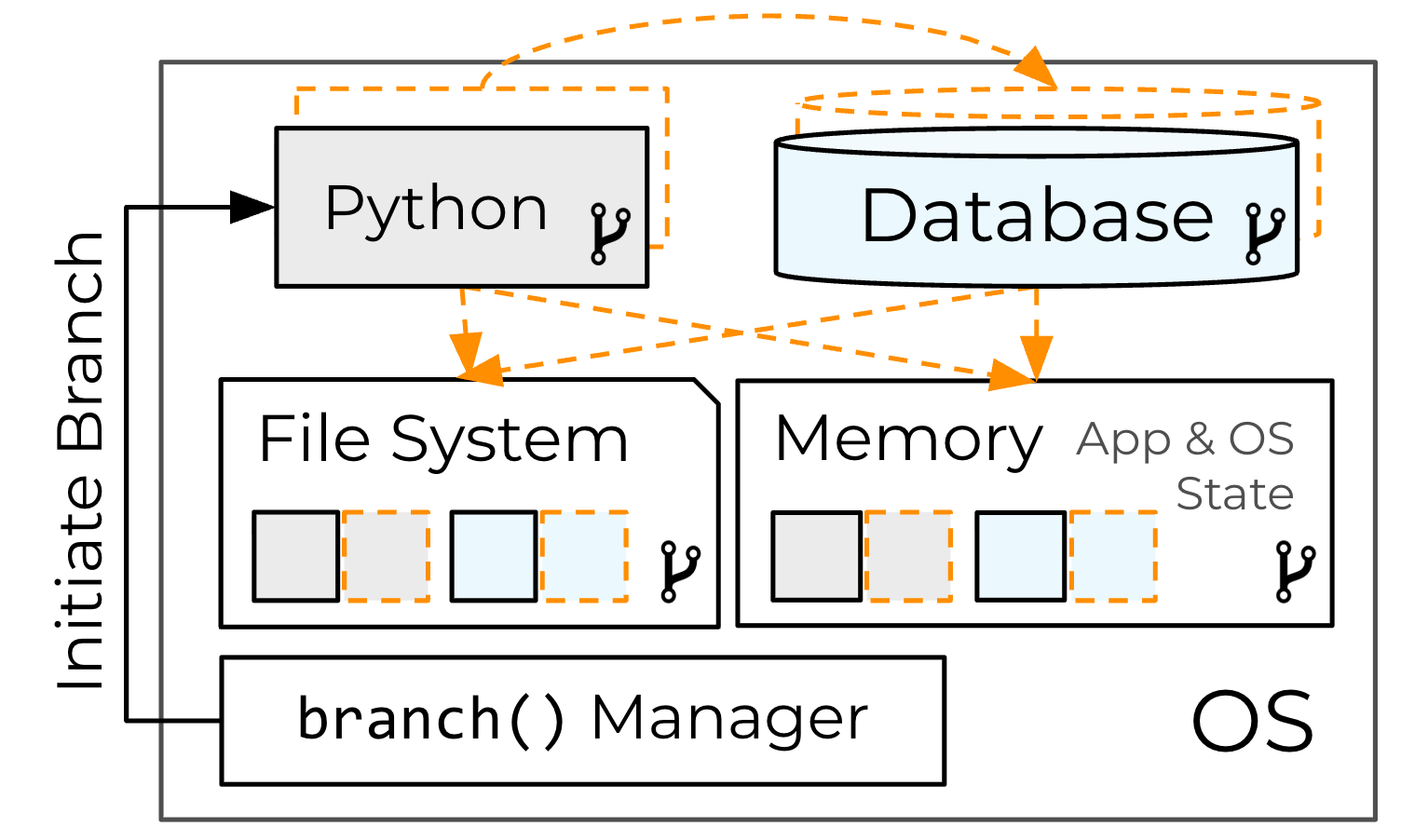}
    \vspace{-2em}
    \caption{Agentic Data Environments must support full state branching.}
    \label{fig:branch}
    \vspace{-1em}
\end{wrapfigure}

The core challenge is that branches have \orange{\emph{transient dependencies}}.
The correct branch state is the closure of objects a live session depends on, including open files, cached metadata, mutated tables, and shell variables.
In \Cref{fig:branch}, the agent branches a Python process with a live DBMS connection. In addition to branching the process’s memory and the file system (\orange{dashed boxes}), the DBMS state referenced by the connection must also be branched (\orange{dashed arrows}) so the agent interacts with an isolated database within the new process branch.
Branching too little state leads to inconsistent or corrupt state, while copying everything is too expensive.
A practical environment must be dependency-aware, and use component-provided branching when possible and OS-level checkpoints otherwise.  

\begin{example}\it
In \Cref{fig:branch}, the agent loads data, installs packages, and inspects a database schema in Python. It then explores alternative schemas.  Only branching the database leaves Python with stale cached metadata, while only branching Python causes speculative database updates to leak across branches.
A correct branch must therefore capture a coherent slice of state across both systems.
\end{example}
The natural fallback primitive is \emph{full OS branching} of the environment that captures, e.g., process-tree state, open files, terminal context, and the filesystem.
This provides a correctness baseline when applications or components do not implement native semantic branching, or when their branching semantics cannot be safely composed.
Existing mechanisms only approximate this capability.
Containers provide isolation, but restore by rerunning from the base image and lose the session's memory and terminal state.
CRIU checkpoints capture process state but are fragile for interactive sessions and grow with runtime footprint~\cite{criu,xu2025systemsfoundationsagenticexploration}.
Virtual machines preserve more state, but their latency and storage costs are too high for agentic exploration~\cite{xu2025systemsfoundationsagenticexploration}.

We have developed Checkpoint-lite (\texttt{Chkpt}) to support a \emph{state branching} abstraction where full OS branches provide correctness while component-specific branches (e.g., DBMS and filesystem branches) are used when available.
\texttt{Chkpt} avoids 
container-style repackaging and shares unchanged state via copy-on-write.   In our preliminary results, using \texttt{Chkpt} to checkpoint only the file system takes $66\,\mathrm{ms}$ and is independent of its size.  In contrast, using containers takes $11.21\,\mathrm{s}$ to checkpoint 2GB.
To checkpoint $1\,\mathrm{GB}$ of in-memory and file system state,  \texttt{Chkpt} takes $1.46\,\mathrm{s}$ as compared to Podman with CRIU~\cite{criu-podman}, which takes $8.84\,\mathrm{s}$.

\subsubsection{Research Directions} 
Ultimately, a data environment must be \emph{agentic}: it should interact with the executing agent to determine what state, consistency, and fidelity are sufficient for a task, trading off correctness, cost, and performance during branching.
This requires branching across heterogeneous applications and system services, inferring the minimal cross-component state induced by transient dependencies, and capturing consistent snapshots across components (e.g., DBMS, filesystem, Python) that lack a shared transactional boundary.
External services (e.g., web APIs) must expose versioned interfaces or be approximated through replay or world models~\cite{hao2023reasoning,allen1983planning}.
Efficient exploration further requires managing large trees of short-lived branches through copy-on-write sharing, deduplication, and garbage collection.

\subsection{Data Flow Control for Data Safety}

While exploration addresses \emph{state safety}, it does not preserve \emph{data safety}.  
Although traditional databases can control \emph{who} may access data, agents with legitimate access can corrupt records, combine data in ways that violate policies, or insert hallucinations that propagate through the environment.
What matters is to constrain \emph{how} data may be derived and used. 
We refer to this capability as \textbf{\textit{Data Flow Control}} (DFC). Conceptually, DFC policies define permitted flows of information from sources (relations, files, RAG stores) to sinks (tables, files, prompts, tools, agent memory, or external APIs).
\begin{example}\it
A tax preparation agent has access to read a database of credit card receipts to determine which can be deducted as business expenses.  
For instance, the agent executes 
\code{Q = INSERT INTO Expenses SELECT id, item, cost, est\_deduction(*) FROM Receipts;}
Accounting is a heavily regulated industry, and violating a number of data use policies can lead to repercussions in finances, criminality, and reputation.  Thus, the query must also be:
\begin{itemize}[itemsep=0em]
\item \textbf{Private:} One user's \code{Receipts} must never be released; they must always be aggregated across  users.  
\item \textbf{Grounded:}  Agents may hallucinate and insert non-existent receipts into \code{Expenses}, thus every row inserted  into \code{Expenses} must be derived from a receipt.  
\item \textbf{Law-abiding:}  Deductions must comply with tax regulations; for example, no more than 50\% of a meal may be deducted as a business expenses~\cite{irs2024mealdeduction}.
\end{itemize}
A query may be syntactically and semantically correct but still violate these constraints.  For instance, it may  return raw receipts in a report (privacy violation),  insert a non-existent Porsche purchase into \code{Expenses} (grounding violation),  or expense the full amount of a steak dinner (law violation). Unfortunately, existing safety and database mechanisms cannot guarantee compliance with any of these policies.
%
\end{example}
Similar policies arise across regulatory requirements, multi-tenant isolation, and prompt injection. They all serve to  restrict \emph{data derivations}. Although access control governs data access, integrity constraints govern stored data, and provenance explains output derivations, none enforce {\it allowable} derivations during execution.

A popular strategy encodes policies in prompts or uses LLMs to evaluate whether a query is safe \cite{zhang2024shieldllm}. These methods are inherently probabilistic, provide no formal guarantees, and degrade as policy complexity and data scale grows.  
For instance, we used frontier models to check whether a query violates a trivial policy (e.g., {\it average \code{qty} less than 30}) by including the query, policy, and the first 100 query results. On 13 TPC-H queries, GPT-5.2 and Claude Opus 4.6 take $0.8-2.2$ seconds, $0.11-0.365\textcent$, and only achieve an F1 measure of $0.4$.

\subsubsection{Enforcing Data Flow Control in the DBMS}
Within the DBMS, relational provenance already describes how output tuples are derived from input records~\cite{Green2007ProvenanceS}. A natural starting point are DFC policies that are logical predicates over the contributing input tuples for each result.
Our work finds that policy semantics must be \emph{optimizer-invariant}~\cite{Summers2025PleaseDK}: query optimizers rewrite execution plans, which changes the structure of provenance expressions. Policies must therefore depend only on the contributing input tuples (provenance monomials), not the physical execution plan.   
DFC policies can reference arbitrary relations and system context, and may optionally call external functions such as LLMs for semantic checks, though enforcement remains deterministic.
For example, a tax compliance policy may limit meal deductions in an expense report:
\begin{quote}\small
\texttt{SOURCE Receipts SINK Expenses \\
CONSTRAINT Expenses.biz\_use <= 0.5 OR Receipts.category != 'Meal'}
\end{quote}
Although policies are logically over provenance, physical enforcement must not materialize provenance, as that can slow queries by over $10{,}000\times$.
Instead, a lightweight rewrite layer compiles policies to execute as part of the base query.  This rewrite-based approach is thus portable to DBMS engines without modifying their internals.

Across five engines (DuckDB, Umbra, PostgreSQL, DataFusion, and SQL Server), we show that enforcing DFC policies incurs ${\sim}0$ overhead on TPC-H queries and provides deterministic guarantees. 
These results show that logical data-flow policies can and should be enforced inside the query engine. 

\subsubsection{Data Flow Control Beyond the DBMS}

The DBMS is only one component of an agentic workflow. In practice, agents move data across many systems: querying databases, processing results in Python, writing files, constructing prompts, calling external APIs, and committing outputs back to persistent storage. Once data leaves the query processor, relational provenance alone is insufficient because the derivation now spans multiple tools and representations.
\begin{example}\it
Consider again the tax-report agent. It issues two aggregate queries over the \texttt{Receipts} table: one computes total travel reimbursement for a department, and the other computes the same total but without a specific employee. Although each query individually satisfies an aggregation policy, the answers together reveal that employee's expense amount. The relevant question is therefore not whether a single query is allowed, but whether the entire cross-tool derivation is permissible.
\end{example}
Enforcing DFC in this setting requires tracking how information flows across the entire agentic workflow rather than within a single query. Similar to optimizer invariance, policies should depend on the underlying information being propagated, and {\it not} the representation used to carry it. Under {\it Representation Invariance}, the same policy should apply irrespective of whether the data moves through SQL, Python, files, or prompts.

Prior work on \emph{Transparent Computing}~\cite{bates2015trustworthy, pasquier2017practical}  argued for observable system behavior across the compute stack. However, they focus at the byte and syscall level to track how data moves between processes or files, but not how {\it information} propagates. Enforcing DFC requires raising the semantic level to track the flow of records, aggregates, summaries, and other derived artifacts across tools and representations.

Agentic data environments already expose natural enforcement boundaries---tool invocations, prompt construction, memory updates, file writes, and network calls. These events provide the points where provenance and policy labels can propagate, where constraints can be enforced before data reaches a sink, and rich semantics that allows for pushing policies into execution similar to within the DBMS.

\subsubsection{Research Directions.}
The long-term goal is to extend DFC from per-query enforcement to environment-level guarantees over end-to-end agent workflows that interact with databases, processes, files, and other agents. Beyond a more expressive policy language, this requires tracking fine-grained data flows across heterogeneous components and propagating annotations through semantic transformations such as summarization or classification. At the same time, physical enforcement must remain dynamic and lightweight: agents synthesize workflows online, so enforcement must operate incrementally and push checks down the execution stack to avoid costly materialization.

Policies may be authored by teams, organizations, regulatory bodies, or end users, and may scale to thousands or millions of rules. When combined with agent exploration, simply rejecting an action due to a policy violation produces a sparse and uninformative signal. Instead, the data environment should actively guide agents by explaining relevant policies, identifying the causes of violations, suggesting safe alternatives, and providing contextual feedback that helps agents revise their plans. In conjunction with branching, such feedback allows agents to explore policy-compliant alternatives while maintaining strong safety guarantees.

\section{Putting the Pieces Together}\label{s:conclusions}
As the introduction argued, agentic automation must \emph{increase the benefits of automation} and \emph{bound the consequences of failure}---each addressed by the preceding sections. 
AIM, AIR, and ADE in  \Cref{s:benefits} increase benefits by finding, collecting, expanding, and refining the information that agents reasons over.  
Branching and Data Flow Control in \Cref{s:costs} bound the costs of failure by empowering agents to explore alternatives without corrupting shared state, and deterministically constraining how data is accessed, combined, and released. These ensure that agent autonomy does not lead to unbounded risk.

\begin{wrapfigure}{r}{0.30\textwidth}
\centering
\includegraphics[width=.95\linewidth]{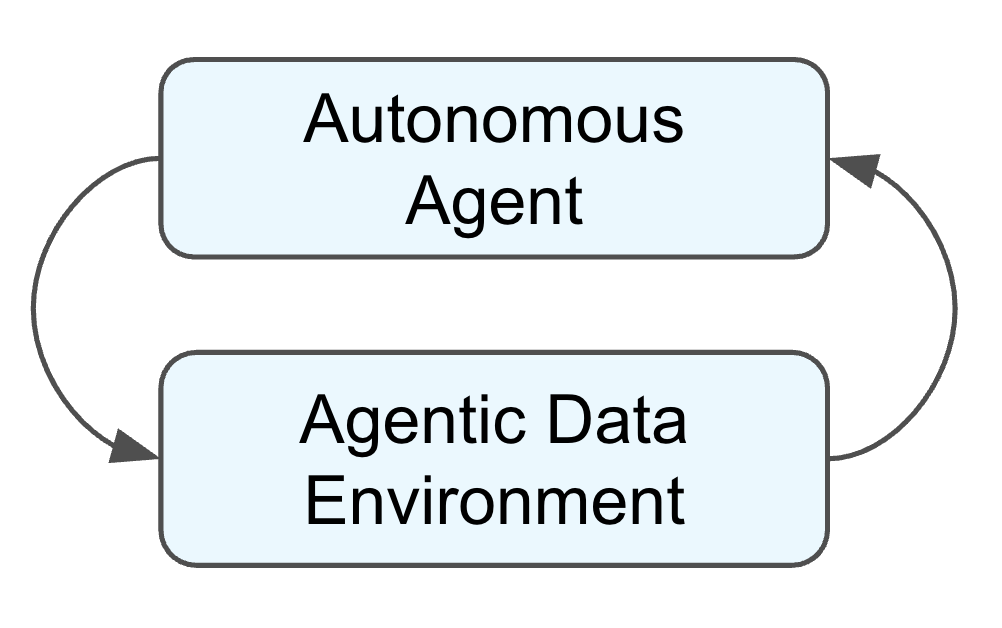}
\vspace{-1em}
\caption{The virtuous agent-environment flywheel.}
\label{fig:shift}
\vspace{-.5em}
\end{wrapfigure}

Automation requires a shift from passive databases to \emph{agentic data environments} to create a virtuous cycle where agents rely on and also improve the environment. Each task produces answers and reusable  artifacts (e.g., schemas, extracted content, indexes, performance models, policies) that make future tasks more accurate, cheaper, and faster.
Over time, these benefits compound, as the goal is not only better models, agents, or datasets, but a better data environment. Agentic data environments therefore evolve the representations, artifacts, and control plan through which agents operate.

While we have mainly focused on inference-time support, agentic data environments also support agent training. Modern post-training paradigms---particularly RL-based methods---must also search, explore, and backtrack through reasoning trajectories. Branching naturally supports this by efficiently materializing and managing persistent intermediate states and accumulating reward signals over time.

\section*{Acknowledgments}
This research was partially supported with funding received from National Science Foundation grants (NSF  1527765, 1564049, 1845638, 1740305, 2008295, 2106197, 2103794, 2312991), an IBM PhD fellowship, as well as corporate support from  Amazon, Google, Adobe, CAIT, Tidalwave, Veris, Shopify, Dandy, Microsoft, Dream Sports, Thinking Machines, Infosys, and Intellect Design. The views and conclusions presented here are those of the authors and should not be interpreted as representing the official positions of the funding organizations.

\bibliographystyle{IEEEtran}
\bibliography{abbreviations, references}

\end{document}